# URM4DMU: AN USER REPRESENTATION MODEL FOR DARKNET MARKETS USERS

*Hongmeng Liu[1], Jiapeng Zhao[1], Yixuan Huo[1], Yuyan Wang[1], Chun Liao[2,3],*
*Liyan Shen[1], Shiyao Cui[2,3], Jinqiao Shi[1]*

[1]Beijing University of Posts and Telecommunications. Beijing, China
[2]Institute of Information Engineering, Chinese Academy of Sciences. Beijing, China
[3]School of Cyber Security, University of Chinese Academy of Sciences. Beijing, China

## ABSTRACT

Darknet markets provide a large platform for trading illicit goods and services due to their anonymity. Learning an invariant representation of each user based on their posts on different markets makes it easy to aggregate user information across different plat forms, which helps identify anonymous users. Traditional user representation methods mainly rely on modeling the text information of posts and cannot capture the temporal content and the forum interaction of posts. While recent works mainly use CNN to model the text information of posts, failing to effectively model posts whose length changes frequently in an episode. To address the above problems, we propose a model named URM4DMU(User Representation Model for Darknet Markets Users) which mainly improves the post representation by augmenting convolutional operators and self-attention with an adaptive gate mechanism. It performs much better when combined with the temporal content and the forum interaction of posts. We demonstrate the effectiveness of URM4DMU on four darknet markets. The average improvements on MRR value and Recall@10 are 22.5% and 25.5% over the state-of-the-art method respectively.

## 1. INTRODUCTION

Darknet markets contain vast resources for the illicit drug trade, adult content and other illicit services [1, 2]. The anonymity of the darknet makes it an ideal environment for criminal discussions [3]. Users on darknet markets are resilient to closures. They will quickly migrate to newer markets when one market shuts down [4]. Learning an in variant representation for each user based on their posts on different markets will link malicious users who use multiple accounts, which makes it convenient to aggregate user information across different platforms. It provides more useful information for the analysis of anonymous users' identities, making user representation learning on these markets become a compelling problem.

Traditional techniques for such tasks rely upon feature engineers [3], that is, using features from text corpora, such as the high frequency words, capitalization, punctuation style, word or character n-grams and function words usage [5, 6], as a user's 'signature'. However, such techniques perform poorly for short text corpora in highly anonymized environments. Convolutional neural networks (CNN) are introduced for modeling the text in user representation model [7, 8, 9]. On this basis, Andrews and Witteveen (2019) [10] proposed to combine the analysis of graph context information and temporal characteristics with text representation for user representation. Pranav Maneriker (2021) [11] developed a novel stylometry based multitask learning approach that leverages graph context to construct low-dimensional representations of short episodes of user activity on darknet markets. They first applied the method to the analysis of dark web data and achieved good results on four darknet market forums datasets. However, the core component of the method mainly relies upon the convolution operation. It has a significant weakness that it only operates on a local neighborhood, thus missing global information [12]. Transformer [13] has emerged as a recent advance to capture long-range interactions and has mostly been applied to sequence modeling. However, since the transformer takes into account all the elements with a weighted averaging operation that disperses the attention distribution, it may overlook the relation of neighboring elements (i.e. n-grams) which are important for modeling text [14, 15, 16].

| Time | User | Post Text and Length |
|---|---|---|
| 2013/7/20 | sh4d3r1950 | Megaupload ?? it was taken down by FBI …file limit . [Length:40] |
| 2013/7/25 | billybob33 | Hello. I have been trying…just tell me so I can remove it. Thanks. [Length:282] |
| 2013/7/26 | sh4d3r1950 | Simple Video Explanation frombitcoinmining.com…was the only way to mine bitcoins…the difficulty rises as well. [Length:1178] |
| 2013/7/31 | sweetbabyjesus | Guten tag! …nitrocellulose…nitroglycerin Ingredients…be made to pellet form? [Length:322] |
| 2013/9/25 | sh4d3r1950 | who cares FBI. [Length:3] |
| 2013/10/17 | billybob33 | That is why I fe'd …a review for your product here on the forum. [Length:93] |

Fig. 1. An example with the length of a user's posts which changes in a large range frequently. The content form of darknet markets is similar to the forum.

As shown in Fig 1, the length of a user's posts may change in a large range frequently, which requires the model to be able to not only capture the local features but also understand the long range post content. Therefore, we propose to concatenate convolutional feature maps with a set of feature maps produced via self-attention through an adaptive gate mechanism to address the above challenges. In summary, our main contributions are as follows:

First, we proposed a user representation model for darknet markets users URM4DMU. It takes advantage of the CNN in capturing local features and the transformers in modeling long sequential information, and combines the temporal content and forum interactions of the user. It is more adaptive to model posts with various lengths of the same user in comparison with traditional methods.

Second, we reveal the importance and remarkable effect of combining the temporal content and forum interactions with our text representation method for modeling darknet markets users.

Third, we demonstrate the effectiveness of URM4DRU on four darknet markets - Black Market Reloaded, Agora Marketplace, Silk Road, and Silk Road 2.0. The average improvements on MRR value and Recall@10 are 22.5% and 25.5% over the state-of-the-art method respectively, which highlights the benefits of URM4DMU.

## 2. METHOD

As shown in Fig 2, URM4DMU 1 contains two key components, namely the post embedding and the episode embedding. The post embedding is derived from the text, time and the structural context of a post. We further use the episode embedding to model users' posts over a period on darknet markets. Each episode e of length Le consists of multiple tuples of texts, times, and contexts $e = \{(t_i, \tau_i, m_i), 1 \leq i \leq L_e\}$. A neural network architecture $f(\theta)$ maps each episode e to combined representation $e \in R^E$. We design a metric learning task to ensure that episodes with the same author have similar representations.

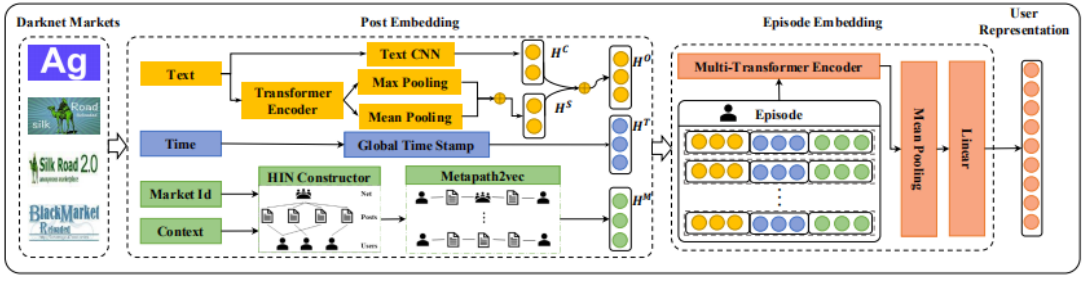

Fig. 2. The framework of the proposed URM4DMU.

## 2.1. Text Embedding

The text embedding is designed to extract semantic features from the post and project the extracted semantic features into the information space. The sentence of the post is first padded to maintain the uniform length $Lp$ of all sentences. Each token of the sentence is mapped to a dt dimensional continuous space by a one-hot coding layer followed by an embedding matrix $Et$ of dimensions $|V| * d_t$ where V is the token vocabulary. The post is represented as a sequence embedding of dimension $L_p \times d_t$ which is concatenated by all the token vectors. Then the model input is $X = [X_0, X_1, ..., X_{L_p-1}]$. After that, we apply the CNN to extract local features, the transformer to model the long sequential information, and a valve component to learn the importance of the above two components for user modeling.

### 2.1.1. Convolutional Over Text

The CNN is to extract local features of the token sequence. The multiple convolution kernels with different sizes (sizes are 2,3,4,5) are set to extract key information and capture local correlations:

$$\mathbf{C} = \text{CNN}(\mathbf{X}), \quad (1)$$

C is fed into a dense layer: $H^C = W^C \cdot C + b^C$. $H^C$ is the output of CNN which involves key local features. The dimensional is $d_e$.

### 2.1.2. Self-attention Over Text

After getting the sequence representation X, the Positional Encoding is introduced to get position information. The vector representation of the token will be obtained by adding both of them. Then, three vectors $q_t$、$k_t$、$v_t$ are obtained according to the embedding vector $x_t (0 \leq t \leq L_p - 1)$. A score for each vector is computed with the following equation: score = $q_t \times k_t$. To stabilize the gradient, score normalization is used. It is divided by $\sqrt{d_k}$. The score value goes through the Softmax activation function to obtain the weight. Once the weight is obtained, it is multiplied by the value vectors of the corresponding tokens $v_t$, and a weighted score for each input vector $v_t$ is achieved. The final output is $z = \sum_{i=0}^{L_p-1} v_t$. The calculation formula is as follows:

$$\text{Attention}(\mathbf{Q}, \mathbf{K}, \mathbf{V}) = \text{Softmax}\left(\frac{\mathbf{QK^T}}{\sqrt{d_k}}\right)\mathbf{V}, \quad (2)$$

Multi-Head Attention (MHA) can provide m representation subspaces for attention. In each subspace, different Q、K、V weight matrices are trained, and each matrix is randomly initialized and generated. The token embeddings are then trained to project into different representation subspaces.

$$\text{MultiHead}(\mathbf{Q}, \mathbf{K}, \mathbf{V}) = \text{Concat}(\mathbf{Att}_1, \mathbf{Att}_2, \cdots, \mathbf{Att}_m)W^A, \tag{3}$$

where $\mathbf{Att}_i = \text{Attention}(\mathbf{QW}_i^Q, \mathbf{KW}_i^K, \mathbf{VW}_i^V)$, $1 \leq i \leq m$, and $W^A$ is a learned linear transformation.

Then the latent semantic feature map S is produced through Feed Forward and Add&Norm, is the number of MHA:

$$\mathbf{S} = \pi\text{MHA}(\mathbf{X}), \tag{4}$$

**2.1.3. Valve Component**

To take advantage of the CNN in capturing local features and the transformers in modeling long sequential information, we apply pooling layers and dense layers to project S into the ds dimensional information space:

$$\mathbf{H}_{max}^S = \text{Linear}(\text{MaxPooling}(\mathbf{S})), \tag{5}$$
$$\mathbf{H}_{mean}^S = \text{Linear}(\text{MeanPooling}(\mathbf{S})), \tag{6}$$

We fuse $\mathbf{H}_{max}^S$、$\mathbf{H}_{mean}^S$ and $H^C$ to output a $d_t (d_t = d_c + 2d_s)$ dimensional information enhanced semantic feature map $H^O$ through the AdaGate [17] function:

$$H^O = AdaGate\left(\mathbf{H}_{max}^S, \mathbf{H}_{mean}^S, H^C, \varepsilon\right) = H^C + Valve(H^\lambda, \varepsilon) \odot \left(\mathbf{H}_{max}^S + \mathbf{H}_{mean}^S\right), \tag{7}$$

Where $\odot$ stands for an element-wise product. The values in $H^\lambda$ are in probability form, and the Valve function is designed to restore less-confident entries (with probability near 0.5) for matching with elements in $\mathbf{H}_{max}^S + \mathbf{H}_{mean}^S$. Concretely, for every unit $\lambda \in H^\lambda$.

$$Valve(\lambda, \varepsilon) = \begin{cases} \lambda, & 0.5 - \varepsilon \leq \lambda \leq 0.5 + \varepsilon \\ 0, & otherwise \end{cases} \tag{8}$$

where $\varepsilon$ is an empirical hyper-parameter which is used for tuning the threshold of confidence. Specifically, we dump all global information if $\varepsilon = 0$, and accept all global information if $\varepsilon = 0.5$. Therefore, the element-wise production exploits Valve as a filter that only extracts necessary information.

**2.2. Time Embedding**

The temporal content plays a key role in modeling the user [10, 11], especially for our model. It contains important information indicating when the post was created and is available at different granularities across darknet market forums. It involves some behavior of users on the darknet market. To obtain a consistent time embedding, we only consider all the date information (date) with the smallest granularity available in the market. The day of the week and date (i.e. day of the year) of each post are used to compute the time embedding by projecting it into a $d\tau$ dimensional vector $H^T$ as a global time stamp. It is fed into the time encoder for time embedding.

**2.3. Structural Context Embedding**

The structural context embedding aims at capturing user behavior on forums, which is first proposed by previous work [11]. A heterogeneous graph is constructed from forum posts by four types of nodes: user (U), sub-forum (S), thread (T), and post (P). Each edge indicates a post of a new thread (U-T), a reply to an existing post (U-P) or an inclusion (T-P, S-T) relationship. The metapath2vec framework [18] with specific meta-path schemes is designed for darknet forums to capture user behavior. To fully capture the semantic relationships in the heterogeneous graph, all meta-paths starting from and ending at a user node are considered. Seven meta-path schemes:

UPTSTPU, UTSTPU, UPTSTU, UTSTU, UPTPU, UPTU, and UTPU are designed to capture user behavior. A final $dm$ dimensional embedding $H^M$ is generated for the context of a post. The learned embeddings will preserve the semantic relationships between each subforum, including posts and relevant users.

**2.4. Episode Embedding**

The embeddings of each component of a post in an episode are concatenated into a $de = dt + d\tau + dm$ dimensional embedding. An episode with L posts has a $L * de$ embeddings. We follow the architecture proposed by Andrews and Bishop (2019) [10].

After episode embedding, we can use a final metric learning loss corresponding to the task-specific $g(\varphi)$, and then train the parameters $\theta$ and $\varphi$. The framework, as mentioned above, results in a train able model referred to as Single-Task Learning for a single market Market$_i$. Note that the first half of the framework (i.e., $f(\theta)$) is sufficient to generate embeddings for episodes, making the module invariant to the choice of $g(\varphi)$. However, the embedding modules learned from these embeddings may not be compatible with comparisons across different markets, which motivates our multi-task setup.

Multi-Task Learning. Same as the architecture proposed by Maneriker et al. (2021) [11], we use the cross-dataset to help train the model by combining four different datasets. Since then, the task specific metric learning layer $g(\varphi^{Task_i})$ is selected and a task-specific loss is backpropagated through the network. Note that in the cross dataset, new labels are defined based on whether different usernames correspond to the same author, and episodes are sampled from the corresponding markets. The overall loss function is the sum of the losses across the markets.

**2.5. The Loss Function**

We use the username as a label for the episode within the market and denote each username as a unique label. To train the user embedding function $f(\theta)$, we compose it with a discriminative classifier $g_\varphi: R^D \to R^Y$ with parameters $\varphi$ to predict the author of an episode, where Y is the number of authors in the training set. We follow the architecture proposed by Maneriker et al. (2021) [11] and use Softmax as the loss function.

# 3. EXPERIMENTS

**3.1. Experiment Setup**

**Dataset and Metrics.** Munksgaard and Demant (2016) [19] studied the politics of darknet markets using structured topic models on the forum posts across six large markets. Maneriker et al. (2021) [11] focus on four of the six markets - Silk Road (SR), Silk Road 2.0 (SR2), Agora Marketplace (Agora), and Black Market Reloaded (BMR) (with summary statistics in Table 2).

| Market | Train Posts | Test Posts | Users train | Users test |
|---|---|---|---|---|
| Agora | 175978 | 179482 | 3115 | 4209 |
| SR | 379382 | 381959 | 6585 | 8865 |
| SR2 | 373905 | 380779 | 5346 | 6580 |
| Bmr | 30083 | 30474 | 855 | 931 |

Table 1. Dataset Statistics for Darknet Markets.

**Implementation Details.** We evaluated our method using retrieval based metrics over the representations generated by each approach. We denote the set of all episode representations as

$E = \{e_1, e_2, ..., e_n\}$ and $Q = \{q_1, q_2, ..., q_K\} \in E$ is the sampled subset. We computed the cosine similarity of the query episode representations with all episodes. Let $R_i = \{r_{i1}, r_{i2}, ..., r_{in}\}$ denoted the list of episodes in E ordered by their cosine similarity with episode qi (excluding itself). These settings mainly follow previous work [11]. The following measures are computed.

**Mean Reciprocal Rank:** (MRR) The RR for an episode is the reciprocal rank of the first element (by similarity) with the same author. MRR is the mean of reciprocal ranks for a sample of episodes.

$$MRR(Q) = \frac{1}{k}\sum_{i=1}^{k} \frac{1}{\min_j \left(A(r_{ij}) = A(e_i)\right)} \tag{9}$$

**Recall@k:** (R@k) Following Andrews and Bishop (2019) [10], we define the R@k for an episode $e_i$ to be an indicator denoting whether an episode by the same author occurs within the subset $<r_{i1}, r_{i2}, ..., r_{in}>$. R@k denotes the mean of these recall values over all the query samples.

### 3.2. Baselines

We compare our URM4DMU against three kinds of baselines. First, the Short Text Authorship Attribution Model: it models each post with text classification models, such as TextCNN [9] and Transformer [13]. Second, we use Single-Task Learning. One is IUR [10], which only considers one dataset at a time. The other one is SYSML [11], a representation learning approach that couples temporal content. Finally, we use Multi-Task Learning, a frame work for training the proposed models in a multitask setting across multiple darknet markets, as described in 2.4.

### 3.3. Main Results

Table 2 shows the results on the four datasets. The average improvements on MRR value and Recall@10 are 22.5% and 25.5% over the state-of-the-art method respectively. It shows that the MRR value of URM4DMU outperforms SYSML by 12.8% and 27.8% at most under the single-task and multiple-task settings on Agora dataset respectively. The improvement gap between the single-task setting and the multiple-task setting confirms that the transformer has stronger learning ability for a large amount of data. The improvement of BMR is not as good as the other three, since the data scale of BMR is about one-tenth of the others.

| Method | Agora | | SR | | SR2 | | BMR | |
|---|---|---|---|---|---|---|---|---|
| | MRR | Recall@10 | MRR | Recall@10 | MRR | Recall@10 | MRR | Recall@10 |
| TextCNN [9](only text) | 0.126 | 0.214 | 0.082 | 0.131 | 0.036 | 0.073 | 0.07 | 0.165 |
| Transformer [13](only text) | 0.091 | 0.133 | 0.026 | 0.043 | 0.053 | 0.083 | 0.052 | 0.114 |
| IUR [10](mean pooling) | 0.114 | 0.218 | 0.126 | 0.223 | 0.109 | 0.19 | 0.223 | 0.408 |
| IUR [10](Transformer pooling) | 0.127 | 0.234 | 0.13 | 0.229 | 0.118 | 0.204 | 0.283 | 0.477 |
| SYSML [11](single) | 0.152 | 0.279 | 0.157 | 0.252 | 0.123 | 0.210 | 0.320 | 0.533 |
| SYSML [11](multitask) | 0.303 | 0.466 | 0.227 | 0.363 | 0.304 | 0.464 | 0.438 | 0.642 |
| URM4DMU(single) | **0.280** | **0.464** | **0.254** | **0.41** | **0.250** | **0.425** | **0.342** | **0.547** |
| URM4DMU(multitask) | **0.581** | **0.794** | **0.489** | **0.672** | **0.552** | **0.758** | **0.548** | **0.730** |

Table 2. The best results for each feature group are shown in bold. Variations of the full model are annotated in parentheses. We provide a comprehensive evaluation of URM4DMU across four different darknet forums. Results show that the average improvements on MRR value and Recall@10 are 22.5% and 25.5% over the state-of-the-art method respectively.

### 3.4. Ablation Study

We perform ablation studies to investigate the contributions of specific components of URM4DMU. Due to space limitations, we only present results on the two datasets. As described above, the graph context component models the forum interactions information, and the time component models the temporal content information. We conduct ablation studies on the graph context component and the time component respectively. Results in Table 3 indicate that the time and the graph context are integral components of the user representation. Besides, adding the time and the graph context or not, URM4DMU always outperforms SYSML.

| Method | SR2 MRR | SR2 Recall@10 | BMR MRR | BMR Recall@10 |
|---|---|---|---|---|
| SYSML(single) | 0.123 | 0.210 | 0.320 | 0.533 |
| - graph context | 0.049 | 0.094 | 0.265 | 0.454 |
| - graph context - time | 0.04 | 0.08 | 0.277 | 0.477 |
| URM4DMU(single) | 0.250 | 0.425 | 0.281 | 0.459 |
| - graph context | 0.191 | 0.34 | 0.172 | 0.324 |
| - graph context - time | 0.08 | 0.133 | 0.283 | 0.486 |
| SYSML(multitask) | 0.304 | 0.464 | 0.438 | 0.642 |
| - graph context | 0.214 | 0.347 | 0.396 | 0.602 |
| - graph context - time | 0.167 | 0.28 | 0.366 | 0.575 |
| URM4DMU(multitask) | **0.552** | **0.758** | **0.537** | **0.730** |
| - graph context | 0.537 | 0.742 | 0.515 | 0.706 |
| - graph context - time | 0.365 | 0.507 | 0.489 | 0.672 |

Table 3. Ablation study on URM4DMU. MRR and Recall@10 scores are reported on the test sets of SR2 and BMR datasets.

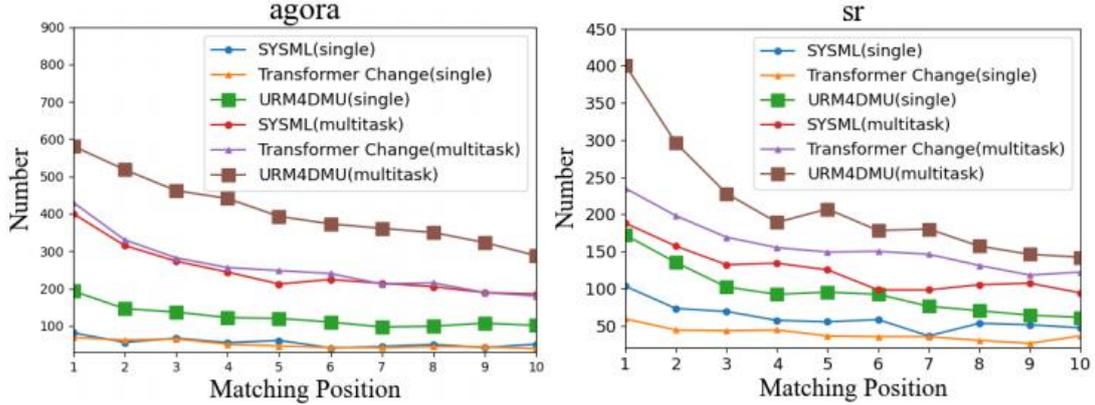

Fig. 3. The comparison between different methods on the number of correct matching positions. The x-coordinate represents the matching of users at the x-th position, and the y coordinate represents the number of correctly matched users.

Figure 3 shows the user alignment results with the representations of SYSML and URM4DMU. We could see that URM4DMU shows obvious advantages over CNN-based models and Transformer based models. The reason is that URM4DMU can not only capture local n-gram features and global dependencies effectively but also preserve sequential information. It also reveals the importance and remarkable effect of combining the temporal content and forum interactions with URM4DMU for modeling darknet markets users.

## 4. CONCLUSION

In this paper, we propose a model named URM4DMU which addresses the problem of learning a representation of each user based on their posts on different markets. The core contribution is the effective use of Transformer in modeling a user's posts whose length change in a large range frequently and in particular, coupled with the temporal content and forum interactions of the user. The consistent and remarkable improvements on four public darknet markets datasets demonstrate the effectiveness of URM4DMU. By linking users across multiple darknet markets, URM4DMU provides more valuable information for the analysis of anonymous users' identities. We will also extend URM4DMU to other darknet platforms like Telegram [20] and Blockchain [21] to collect more valuable information for the analysis of anonymous users' identities. URM4DMU will shed light on the content and user analysis works related to conversation understanding [22].

# 5. REFERENCES

[1] Mirella Lapata, Phil Blunsom, and Alexander Koller, Eds., Proceedings of the 15th Conference of the European Chapter of the Association for Computational Linguistics: Volume 1, Long Papers, Valencia, Spain, Apr. 2017. Association for Computational Linguistics.

[2] Alex Biryukov, Ivan Pustogarov, Fabrice Thill, and Ralf Philipp Weinmann, "Content and popularity analysis of tor hidden services," in 2014 IEEE 34th International Conference on Distributed Computing Systems Workshops (ICDCSW), 2014, pp. 188–193.

[3] Thanh Nghia Ho and Wee Keong Ng, "Application of stylometry to darkweb forum user identification," in Information and Communications Security: 18th International Conference, ICICS 2016, Singapore, Singapore, November 29 – December 2, 2016, Proceedings, Berlin, Heidelberg, 2016, p. 173–183, Springer-Verlag.

[4] Abeer ElBahrawy, Laura Alessandretti, Leonid Rusnac, Daniel Goldsmith, Alexander Teytelboym, and Andrea Baronchelli, 'Collective dynamics of dark web marketplaces," Scientific Reports, vol. 10, 2020.

[5] Julien Hay, Bich-Lien Doan, Fabrice Popineau, and Ouassim Ait Elhara, "Representation learning of writing style," in Proceedings of the Sixth Workshop on Noisy User-generated Text (W-NUT 2020), Online, Nov. 2020, pp. 232–243, Association for Computational Linguistics.

[6] Nacer Eddine Benzebouchi, Nabiha Azizi, Nacer Eddine Hammami, Didier Schwab, Mohammed Chiheb Eddine Khelaifia, and Monther Aldwairi, "Authors' writing styles-based authorship identification system using the text representation vector," in 2019 16th International Multi-Conference on Systems, Signals Devices (SSD), 2019, pp. 371–376.

[7] Yoon Kim, "Convolutional neural networks for sentence classification," in Proceedings of the 2014 Conference on Empirical Methods in Natural Language Processing, EMNLP 2014, October 25-29, 2014, Doha, Qatar, A meeting of SIGDAT, a Special Interest Group of the ACL, Alessandro Moschitti, Bo Pang, and Walter Daelemans, Eds. 2014, pp. 1746–1751, ACL.

[8] Sebastian Ruder, Parsa Ghaffari, and John G. Breslin, "Character-level and multi-channel convolutional neural networks for large-scale authorship attribution," ArXiv, vol. abs/1609.06686, 2016.

[9] Prasha Shrestha, Sebastian Sierra, Fabio González, Manuel Montes, Paolo Rosso, and Thamar Solorio, "Convolutional neural networks for authorship attribution of short texts," in Proceedings of the 15th Conference of the European Chapter of the Association for Computational Linguistics: Volume 2, Short Papers, Valencia, Spain, Apr. 2017, pp. 669–674, Association for Computational Linguistics.

[10] Nicholas Andrews and Marcus Bishop, "Learning invariant representations of social media users," in Proceedings of the 2019 Conference on Empirical Methods in Natural Language Processing and the 9th International Joint Conference on Natural Language Processing (EMNLP-IJCNLP), Hong Kong, China, Nov. 2019, pp. 1684–1695, Association for Computational Linguistics.

[11] Pranav Maneriker, Yuntian He, and Srinivasan Parthasarathy, "SYSML: StYlometry with Structure and Multitask Learning: Implications for Darknet forum migrant analysis," in Proceedings of the 2021 Conference on Empirical Methods in Natural Language Processing, Online and Punta Cana, Dominican Republic, Nov. 2021, pp. 6844–6857, Association for Computational Linguistics.

[12] Irwan Bello, Barret Zoph, Ashish Vaswani, Jonathon Shlens, and Quoc V. Le, "Attention augmented convolutional networks," 2019 IEEE/CVF International Conference on Computer Vision (ICCV), pp. 3285–3294, 2019.

[13] Ashish Vaswani, Noam Shazeer, Niki Parmar, Jakob Uszkoreit, Llion Jones, Aidan N Gomez, Lukasz Kaiser, and Illia Polosukhin, "Attention is all you need," Advances in neural information processing systems, vol. 30, 2017.

[14] Baosong Yang, Zhaopeng Tu, Derek F. Wong, Fandong Meng, Lidia S. Chao, and Tong Zhang, "Modeling localness for self-attention networks," in Proceedings of the 2018 Conference on Empirical Methods in Natural Language Processing, Brussels, Belgium, Oct.-Nov. 2018, pp. 4449–4458, Association for Computational Linguistics.

[15] Baosong Yang, Longyue Wang, Derek F. Wong, Lidia S. Chao, and Zhaopeng Tu, "Convolutional self-attention networks," in Proceedings of the 2019 Conference of the North American Chapter of the Association for Computational Linguistics: Human Language Technologies, Volume 1 (Long and Short Papers), Minneapolis, Minnesota, June 2019, pp. 4040–4045, As sociation for Computational Linguistics.

[16] Maosheng Guo, Yu Zhang, and Ting Liu, "Gaussian transformer: A lightweight approach for natural language inference," in Proceedings of the Thirty-Third AAAI Conference on Artificial Intelligence and Thirty-First Innovative Applications of Artificial Intelligence Conference and Ninth AAAI Symposium on Educational Advances in Artificial Intelligence. 2019, AAAI'19/IAAI'19/EAAI'19, AAAI Press.

[17] Xianming Li, Zongxi Li, Haoran Xie, and Qing Li, "Merging statistical feature via adaptive gate for improved text classifi- cation," in AAAI, 2021.

[18] Yuxiao Dong, Nitesh V. Chawla, and Ananthram Swami, "metapath2vec: Scalable representation learning for heterogeneous networks," in Knowledge Discovery and Data Mining,
2017.

[19] Rasmus Munksgaard and Jakob Johan Demant, "Mixing politics and crime - the prevalence and decline of political discourse on the cryptomarket.," The International journal on drug policy, vol. 35, pp. 77–83, 2016.

[20] Jason Baumgartner, Savvas Zannettou, Megan Squire, and Jeremy Blackburn, "The pushshift telegram dataset," in Proceedings of the International AAAI Conference on Web and Social Media, 2020, vol. 14, pp. 840–847.

[21] Zibin Zheng, Shaoan Xie, Hong-Ning Dai, Xiangping Chen, and Huaimin Wang, "Blockchain challenges and opportunities: A survey," International journal of web and grid services, vol. 14, no. 4, pp. 352–375, 2018.

[22] Zhuosheng Zhang, Yuwei Wu, Hai Zhao, Zuchao Li, Shuailiang Zhang, Xi Zhou, and Xiang Zhou, "Semantics-aware bert for language understanding," in Proceedings of the AAAI Conference on Artificial Intelligence, 2020, vol. 34, pp. 9628–9635.